\documentclass[11pt]{article}

\usepackage[final]{acl}

\usepackage{times}
\usepackage{latexsym}
\usepackage{color}

\usepackage[T1]{fontenc}

\usepackage[utf8]{inputenc}

\usepackage{microtype}

\usepackage{inconsolata}

\usepackage{graphicx}

\usepackage[utf8]{inputenc} 
\usepackage[T1]{fontenc}    
\usepackage{hyperref}       
\usepackage{url}            
\usepackage{booktabs}       
\usepackage{amsfonts}       
\usepackage{nicefrac}       
\usepackage{microtype}      
\usepackage{colortbl}
\usepackage{array}

\usepackage{algorithm}
\usepackage{algorithmic}
\usepackage[utf8]{inputenc} 
\usepackage[T1]{fontenc}    
\usepackage{url}            
\usepackage{booktabs}       
\usepackage{amsfonts}       
\usepackage{nicefrac}       
\usepackage{microtype}      
\usepackage{bm}
\usepackage{enumitem}
\usepackage{graphicx}
\usepackage{amsmath}
\usepackage{subfigure}
\usepackage{amssymb}
\usepackage{caption}
\usepackage{longtable}
\usepackage{soul}
\usepackage{color}
\usepackage{appendix}
\usepackage{comment}
\usepackage{xcolor}
\usepackage{tcolorbox}
\tcbuselibrary{skins}

\usepackage{tikz}
\usetikzlibrary{trees}
\usetikzlibrary{positioning,fit}

\usepackage{forest}
\usepackage{array}
\usepackage{xcolor} 
\usepackage{multirow} 
\usepackage{tabularx} 

\usepackage{multirow}
\usepackage{makecell}
\usepackage{graphicx}

\definecolor{level1}{RGB}{204, 229, 255} 
\definecolor{level2}{RGB}{229, 255, 204} 
\definecolor{level3}{RGB}{255, 229, 204} 

\usepackage{tikz}
\usetikzlibrary{shapes.geometric, arrows.meta, positioning, decorations.pathmorphing, shadows}

\title{From Passive Metric to Active Signal: The Evolving Role of \\ Uncertainty Quantification in Large Language Models}

\author{Jiaxin Zhang$^{1}$\thanks{ Correspondence to \texttt{jiaxin.zhang@salesforce.com}} \quad  Wendi Cui$^2$ \quad Zhuohang Li$^3$, \\ {\bf Lifu Huang$^4$ \quad {\bf Bradley Malin$^{3,5}$ \quad Caiming Xiong$^{1}$ \quad Chien-Sheng Wu$^{1}$}}
\\ $^1$Salesforce AI Research \quad $^2$Intuit \quad $^3$Vanderbilt University \\ \quad $^4$ University of California,
Davis \quad $^5$Vanderbilt University Medical Center\\
}

\begin{document}
\maketitle

\begin{abstract}
While Large Language Models (LLMs) show remarkable capabilities, their unreliability remains a critical barrier to deployment in high-stakes domains. This survey charts a functional evolution in addressing this challenge: the evolution of uncertainty from a passive diagnostic metric to an active control signal guiding real-time model behavior. We demonstrate how uncertainty is leveraged as an active control signal across three frontiers: in \textbf{advanced reasoning} to optimize computation and trigger self-correction; in \textbf{autonomous agents} to govern metacognitive decisions about tool use and information seeking; and in \textbf{reinforcement learning} to mitigate reward hacking and enable self-improvement via intrinsic rewards. By grounding these advancements in emerging theoretical frameworks like Bayesian methods and Conformal Prediction, we provide a unified perspective on this transformative trend. This survey provides a comprehensive overview, critical analysis, and practical design patterns, arguing that mastering the new trend of uncertainty is essential for building the next generation of scalable, reliable, and trustworthy AI.
\end{abstract}

\section{Introduction}

LLMs have demonstrated unprecedented capabilities across a wide range of natural language tasks, marking a milestone in AI. Yet their inherent unreliability, which manifests through factual errors, biases, and hallucinations, remains a critical barrier to deployment in high-stakes domains such as medicine, law, and finance \citep{bommasani2021opportunities,farquhar2024detecting}. To address this issue, Uncertainty Quantification (UQ) has emerged as a key technology for enhancing trustworthiness.  
Traditionally, UQ has focused on the \textbf{post-hoc evaluation} and calibration of outputs \cite{zhang2021modern,xiong2023can}. Methods based on Bayesian inference, ensembles, or information-theoretic metrics aim to provide confidence scores for single-turn generations, effectively measuring ``how much the model knows'' about its own response \citep{gawlikowski2023survey}. While foundational, this function treats uncertainty as a \textbf{passive, diagnostic metric} attached to completed outputs. Yet such an approach is insufficient for the next generation of LLM systems, which involve multi-step reasoning, interactive environments, and alignment with complex human values \cite{kirchhof2025position, zhang2026auq, zhang2026agentic}.  

The importance of this field has spurred a series of excellent surveys. Some organize the landscape around \textbf{uncertainty estimation}, including token-level analysis, consistency checks, semantic clustering, and entropy \citep{xia2025survey, cui2024divide, shorinwa2025survey, kuhn2023semantic, gao2024spuq, zhang2023sac3}. Others adopt \textbf{theory-grounded perspectives}, linking heuristics to Bayesian and information-theoretic principles \citep{huang2024survey}. Broader work has examined \textbf{confidence calibration} \citep{geng2024survey}, while recent efforts have begun to \textbf{rethink the definition and sources} of uncertainty in the LLM lifecycle, categorizing them along new dimensions such as computational cost or reasoning uncertainty \citep{beigi2024rethinking, liu2025uncertainty,li2025uncertainty}. 

While aforementioned resources provide valuable overviews of how uncertainty can be \emph{measured}, this paper complements that body of work by surveying an emerging technological trend: the evolution of uncertainty from a passive metric to an \textbf{active, real-time control signal}. This enables systems that can \textit{``know what they don’t know''} \cite{kadavath2022language,yin2023largelanguagemodelsknow} and take action based on this self-awareness \cite{betley2025tell}.  

Our key contribution is to categorize and analyze research where uncertainty functions as a control mechanism. While prior work focused on \textbf{how to measure} uncertainty, we focus on \textbf{how to use} it, organizing the discussion around three domains where this functional evolution is most evident:  
\begin{itemize}[leftmargin=10pt]
 \item \textbf{Advanced Reasoning:} How uncertainty guides dynamic reasoning strategies, optimizes computational effort, and triggers self-correction.  
 \item \textbf{Autonomous Agents:} How uncertainty drives decisions on tool use, information seeking, and risk management in interactive settings.  
 \item \textbf{RL and Reward Models:} How modeling uncertainty in human preferences and rewards enables more robust alignment and mitigates failure modes like reward hacking.  
\end{itemize}  

By tracing uncertainty’s evolving role from passive evaluation to active control, we provide a comprehensive overview of this emerging frontier and outline the fundamental challenges and future research directions.  

\forestset{
    highlight0/.style={
        for tree={
            fill=gray!10
        }
    },
    highlight3/.style={
        for tree={
            fill=orange!30
        }
    },
    highlight31/.style={
        for tree={
            fill=orange!15
        }
    },
    highlight32/.style={
        for tree={
            fill=orange!5
        }
    },
    highlight4/.style={
        for tree={
            fill=cyan!30
        }
    },
    highlight41/.style={
        for tree={
            fill=cyan!15
        }
    },
    highlight42/.style={
        for tree={
            fill=cyan!5
        }
    },
    highlight5/.style={
        for tree={
            fill=purple!30
        }
    },
    highlight51/.style={
        for tree={
            fill=purple!15
        }
    },
    highlight52/.style={
        for tree={
            fill=purple!5
        }
    },
    highlight6/.style={
        for tree={
            fill=green!15
        }
    },
    highlight61/.style={
        for tree={
            fill=green!5
        }
    },
    highlight62/.style={
        for tree={
            fill=blue!5
        }
    },
    highlight7/.style={
        for tree={
            fill=blue!10
        }
    },
}

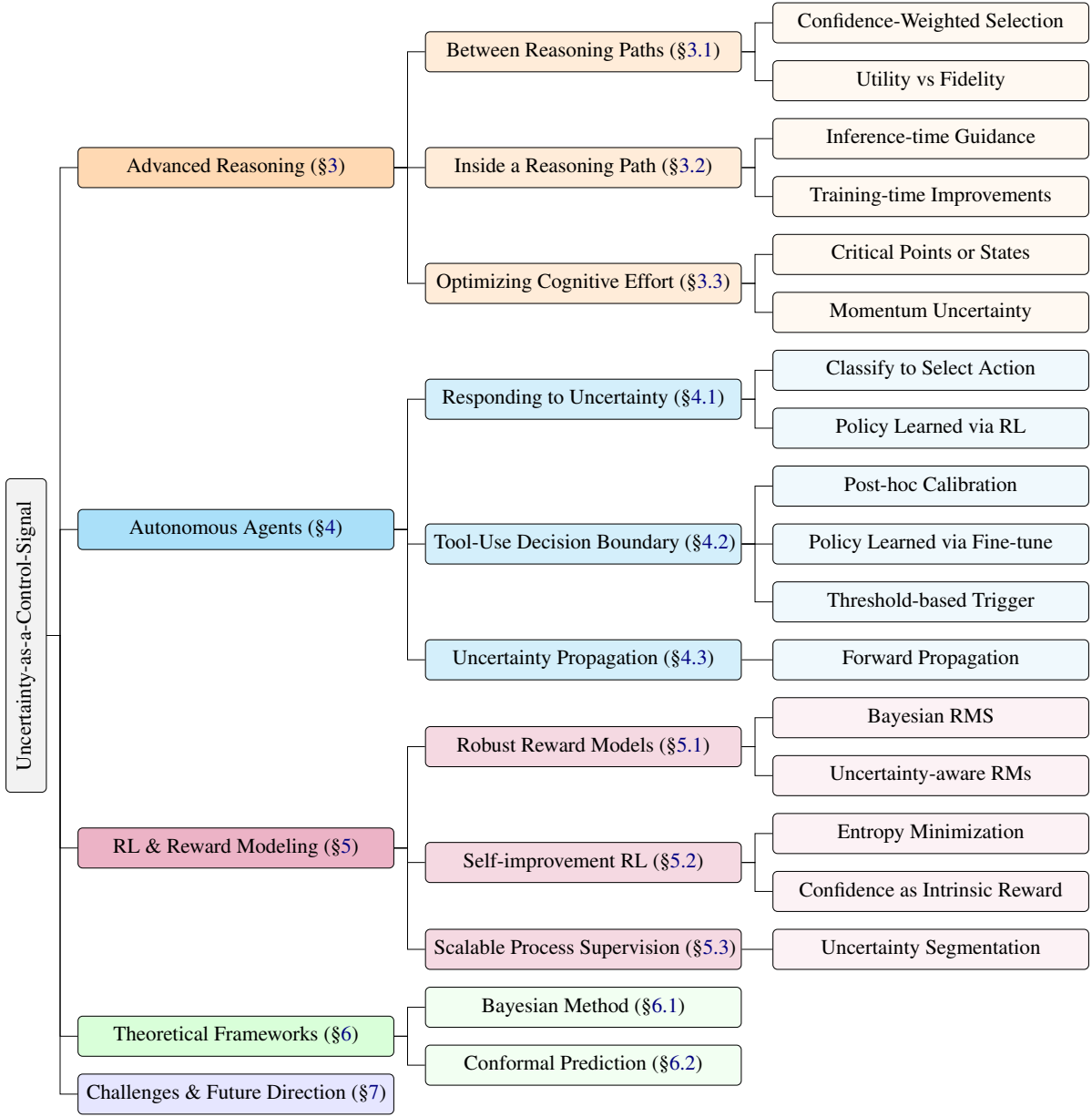
\begin{figure*}[h!]
\centering
\resizebox{0.9\textwidth}{!}{ 
\begin{forest}
    for tree={
        grow=east,
        draw,
        text width=45mm,
        font=\small,
        edge path={
            \noexpand\path [draw, \forestoption{edge}] (!u.parent anchor) -- +(2mm,0) |- (.child anchor)\forestoption{edge label};
        },
        parent anchor=east,
        child anchor=west,
        tier/.wrap pgfmath arg={tier #1}{level()},
        edge={ultra thin},
        rounded corners=2pt,
        align=center,
        text badly centered,
        sibling distance=20mm,
    },
    tikz={
        \definecolor{level1}{RGB}{204, 229, 255}; 
        \definecolor{level2}{RGB}{229, 255, 204}; 
        \definecolor{level3}{RGB}{255, 229, 204}; 
    },
    [Uncertainty-as-a-Control-Signal, rotate=90, child anchor=north, parent anchor=south, highlight0,
        [Challenges \& Future Direction (§\ref{sec:future}), highlight7,
        ],
        [Theoretical Frameworks (§\ref{sec:emerging_frameworks}), highlight6,
             [Conformal Prediction (§\ref{sec:62}), highlight61],
             [Bayesian Method (§\ref{sec:61}), highlight61],
        ],
        [RL \& Reward Modeling (§\ref{sec:alignment}), highlight5,
            [Scalable Process Supervision (§\ref{sec:53}), highlight51,
                [Uncertainty Segmentation, highlight52],
            ],
            [Self-improvement RL (§\ref{sec:52}), highlight51,
                [Confidence as Intrinsic Reward, highlight52],
                [Entropy Minimization, highlight52]
            ],
            [Robust Reward Models (§\ref{sec:51}), highlight51,
                [Uncertainty-aware RMs, highlight52],
                [Bayesian RMS, highlight52]
            ],
        ],
        [Autonomous Agents (§\ref{sec:agents}), highlight4,
            [Uncertainty Propagation (§\ref{sec:43}), highlight41,
                [Forward Propagation, highlight42]],
            [Tool-Use Decision Boundary (§\ref{sec:42}), highlight41,
                [Threshold-based Trigger, highlight42],
                [Policy Learned via Fine-tune, highlight42],
                [Post-hoc Calibration, highlight42]
            ]
            [Responding to Uncertainty (§\ref{sec:41}), highlight41,
                [Policy Learned via RL, highlight42],
                [Classify to Select Action, highlight42]],
        ],
        [Advanced Reasoning (§\ref{sec:reasoning}), highlight3,
            [Optimizing Cognitive Effort (§\ref{sec:33}), highlight31,
                [Momentum Uncertainty, highlight32],
                [Critical Points or States, highlight32],
            ],
            [Inside a Reasoning Path (§\ref{sec:32}), highlight31
                [Training-time Improvements, highlight32],
                [Inference-time Guidance, highlight32]
            ],
            [Between Reasoning Paths (§\ref{sec:31}), highlight31
                [Utility vs Fidelity, highlight32],
                [Confidence-Weighted Selection, highlight32]
            ]
        ]
    ]
\end{forest}
}
\caption{The taxonomy of this survey, illustrating the evolving role of uncertainty to an active control signal across advanced LLM applications, emerging theories and open challenges.}
\label{fig:main_taxonomy}
\end{figure*}

\section{The Limits of Traditional UQ}
\label{sec:limits}

The classical paradigm of UQ provides a foundational, but ultimately limited, framework for assessing the reliability of LLMs. Traditionally, UQ distinguishes between \textit{aleatoric} uncertainty, arising from inherent data noise, and \textit{epistemic} uncertainty, stemming from the model’s lack of knowledge and reducible with more data \citep{kendall2017uncertainties}. The principal objective has been \textbf{post-hoc evaluation}, where a confidence score is assigned after a model generates an output \citep{xia2025survey, shorinwa2025survey,tian2023just}.  

While useful for simple generation tasks, this ``generate-then-evaluate’’ function treats uncertainty as a \textbf{passive, diagnostic metric}. Its inability to provide real-time, actionable feedback becomes especially limiting in the complex, dynamic, and interactive settings that characterize frontier LLM applications \citep{kirchhof2025position}. There are several shortcomings of this strategy:

\begin{itemize}[leftmargin=10pt]
    \item \textbf{Inapplicability to Multi-Step Reasoning:} In chain-of-thought reasoning, early mistakes can derail entire sequences. A final post-hoc score is insufficient; models require continuous uncertainty signals at intermediate steps to backtrack, branch, or adapt in real time.  
    \item \textbf{Insufficiency for Autonomous Agents:} For LLM agents, uncertainty informs various decisions, such as whether to rely on parametric knowledge, invoke tools, or seek human input. A single retrospective score on a text output does not support such proactive choices.  
    \item \textbf{Mismatch with Dynamic and Interactive Systems:} Classical UQ assumes static, monolithic outputs. However, modern LLM systems involve branching reasoning paths, environmental interactions, and iterative alignment loops, requiring uncertainty to evolve dynamically alongside system behavior.  
\end{itemize}  

We believe these limitations call for a functional shift. To build robust and reliable systems, uncertainty must move beyond passive assessment and become an active control signal integrated into the model’s operational loop.

\begin{tcolorbox}[
  colback=gray!5!white, 
  colframe=gray!75!black, 
  title=\textbf{The Evolving Role of Uncertainty Quantification: From Passive Metric to Active Signal},
fonttitle=\bfseries\small,
]
\small
  \noindent \textbf{Passive Metric (Post-hoc Diagnosis)}
  \begin{itemize}[leftmargin=0pt,nosep]
    \item \textbf{When:} Assigns a score \textit{after} generation is complete.
    \item \textbf{Role:} Acts as a diagnostic tool (output reliable?).
    \item \textbf{Nature:} Static and external to the generation process.
  \end{itemize}
\vspace{2mm}
  \noindent \textbf{Active Signal (Real-time Control)}
  \begin{itemize}[leftmargin=0pt,nosep]
    \item \textbf{When:} Intervenes \textit{during} generation via a feedback.
    \item \textbf{Role:} Acts as a control mechanism (trigger actions?).
    \item \textbf{Nature:} Dynamic and integrated into the model's operational loop.
  \end{itemize}
\end{tcolorbox}


\section{Advanced Reasoning}
\label{sec:reasoning}

\begin{table*}[ht]
\centering
\small
\resizebox{\textwidth}{!}{%
\begin{tabular}{l|l|ll}
\toprule
\textbf{Core Concepts}                            & \textbf{Strategy Function}                                                        & \textbf{Uncertainty Signal (The ``What'')} & \textbf{Control Mechanism (The ``How'')}  \\ \midrule
\multirow{6}{*}{\textbf{Between Reasoning Paths}}              & CISC \cite{liu2025cisc} & length-normalized probability   & confidence-weighted voting           \\ \cmidrule(l){2-4} 
  & CER \cite{razghandi2025cer}           & step-wise confidence scores                         & intermediate step aggregation            \\ \cmidrule(l){2-4}
  & UAG \citep{yin2024reasoning}                                    & step-wise uncertainty          & adaptive guidance and backtracking                   \\ \cmidrule(l){2-4}
  & Deep Think \citep{fu2025deepthinkconfidence}                                      & confidence scores                         & weighted path selection                 \\ \cmidrule(l){2-4} 
  & Bayesian Meta-Reasoning \citep{yan2025position}                                 & Bayesian inference              & probabilistic path reasoning          \\ \cmidrule(l){2-4} 
  & s1 \citep{muennighoff2025s1}                              & LLM/reward model scores                    & test-time scaling                    \\ \midrule
\multirow{6}{*}{\textbf{Within Reasoning Path}}                & SPOC \citep{madaan2024boosting}                                      & verification uncertainty                         & proposer-verifier alternation                 \\ \cmidrule(l){2-4} 
  & AdaptiveStep \citep{liu2025adaptivestep}                                 & model confidence              & uncertainty-guided segmentation          \\ \cmidrule(l){2-4} 
  & Uncertainty-Sensitive Tuning \citep{liu2024know}                              & abstention signals                    & two-stage training procedure                    \\ \cmidrule(l){2-4}
  & Uncertainty-Aware FT \citep{zheng2024enhancing}                                        & prediction uncertainty    & modified loss function \\ \cmidrule(l){2-4} 
  & BRiTE \citep{zhong2025brite}                                           & reinforcement signals             & bootstrapped thinking process \\  \cmidrule(l){2-4}
  & External Slow-Thinking \citep{gan2025rethinking_external}                                           & probability of correctness             & data filtering and selection \\  \midrule
\multirow{4}{*}{\textbf{Cognitive Effort Optimization}} & UnCert-CoT \citep{li2025uncertainty}                                        & entropy, probability margins    & threshold-based CoT activation \\ \cmidrule(l){2-4} 
  & MUR \citep{yu2025mur}                                           & momentum uncertainty             & thinking budget allocation \\  \cmidrule(l){2-4}
  & THOUGHT-TERMINATOR \citep{pu2025thoughtterminator}                                           & state sufficiency probability             & overthinking mitigation \\  \cmidrule(l){2-4}
  & TokenSkip \citep{xia2025tokenskip}                                           & controllable compression signals             & chain-of-thought compression \\  \bottomrule
\end{tabular}
}
\caption{A comparative analysis of uncertainty-aware reasoning approaches in LLMs. The table details the specific uncertainty signals and control mechanisms used across three main functions: between-path selection, within-path guidance, and cognitive effort optimization.}
\label{tab:reasoning_comparison}
\end{table*}

In advanced reasoning with LLMs, uncertainty has shifted from a passive, post-hoc quality score to an active internal signal that guides decision-making: from arbitrating \textit{between reasoning paths}, to steering trajectories \textit{within individual reasoning path}, and allocating \textit{cognitive effort} efficiently. Table \ref{tab:reasoning_comparison} provides a comparative analysis of these frameworks, detailing for each method the specific uncertainty signal it uses (the ``what'') and the control mechanism through which it acts (the ``how'').
\subsection{Between Reasoning Paths: Weighted Selection.}
\label{sec:31}
Inference-time scaling, where models generate many reasoning traces and then aggregate them, has become a standard strategy for improving robustness \cite{pan2025coat, liu2025can, zhang2025confidence}.  Uncertainty enables nuanced selection between generated reasoning paths to improve overall accuracy.
\paragraph{Confidence-Weighted Selection.}
Recent work moves beyond the “one path, one vote” function by leveraging uncertainty as a weighting signal \cite{yin2024reasoning, fu2025deepthinkconfidence}. Confidence-Informed Self-Consistency (CISC) \cite{liu2025cisc} assigns each reasoning path a holistic confidence score based on its length-normalized probability, which then weights the final vote. Confidence Enhanced Reasoning (CER) \cite{razghandi2025cer} instead evaluates confidence at crucial intermediate steps, aggregating them into a more robust score. Other approaches apply Bayesian inference to select promising paths \cite{yan2025position}, or trained reward models to compute confidence scores \cite{muennighoff2025s1, li2025testtime_preference_optimization}.

\paragraph{Utility vs. Fidelity Trade-off.}  
Weighted methods expose a tension between the \emph{utility} of confidence scores for local decisions and their \emph{fidelity} for global calibration. As \citet{liu2025cisc} show, methods with strong global calibration (confidence aligning with average accuracy) often struggle to distinguish correct from incorrect reasoning paths on a single question. The key factor is \textbf{Within-Question Discrimination (WQD)}, the ability of confidence to separate right from wrong answers given one problem. A sharp, locally discriminative signal, even if globally ``overconfident'', is more useful for path selection. CER embodies this principle by emphasizing confidence at critical reasoning steps, favoring local discrimination over global fidelity \cite{razghandi2025cer}.  

These approaches illustrate a fundamental trade-off. CER’s fine-grained step evaluation improves robustness in long-chain reasoning, but increases implementation complexity. By contrast, CISC’s holistic scoring is simpler but more sensitive to minor, non-critical errors. Both rely on calibrated confidence estimates; when miscalibrated, the weighting mechanism may amplify errors instead of correcting them. More details in Appendix Table \ref{tab:reasoning_analysis}.  

\subsection{Inside a Reasoning Path: Beyond Inference to Training}
\label{sec:32}
Within a reasoning path, uncertainty is not merely a retrospective confidence measure but an active control signal, guiding reasoning during inference and serving as a training objective \cite{da2025understanding}.

\paragraph{Inference-Time Guidance.}  
Uncertainty provides real-time feedback that allows models to adapt their reasoning as it unfolds \cite{wang2025accelerating_specsearch,hu2024uncertainty}. Uncertainty-Aware Adaptive Guidance (UAG) \citep{huang2024when} monitors step-level uncertainty and retracts to low-uncertainty checkpoints when \textbf{reasoning drifts}. Spontaneous Self-Correction (SPOC) \citep{madaan2024boosting} assigns the model dual roles of proposer and verifier, using uncertainty to \textbf{action selection}: continuation, backtracking, or revision. AdaptiveStep \citep{liu2025adaptivestep} aligns reasoning with natural uncertainty-guided boundaries rather than rule-based segmentation, improving supervision and interpretability. In this view, uncertainty shapes both the unfolding of reasoning and the \textbf{structural units} within it \cite{yin2024reasoning}.  

\paragraph{Training-Time Improvements.}  
Uncertainty also drives advances in model training \cite{zhong2025brite}. \textbf{Uncertainty-Sensitive Tuning} \citep{liu2024know} teaches models to abstain under high uncertainty, then restores general capabilities while retaining calibrated restraint. \textbf{Uncertainty-Aware Fine-Tuning} modifies the loss function itself \citep{zheng2024enhancing}, rewarding higher uncertainty on ultimately incorrect predictions to produce more reliable estimates. Other approaches apply \textbf{Uncertainty-guided data filter} to emphasize plausible examples \cite{gan2025rethinking_external}. These methods elevate uncertainty from a secondary signal to a primary learning objective in training.  

In summary, inference-time methods offer immediate correction without retraining, but remain limited by the model’s intrinsic self-correction ability. Training-time approaches incur a higher cost upfront but yield models with fundamentally stronger uncertainty awareness across downstream tasks. 

\begin{table*}[!h]
\centering
\small
\resizebox{\textwidth}{!}{%
\begin{tabular}{l|l|ll}
\toprule
\textbf{Core Concepts}                            & \textbf{Strategy Function}                                                        & \textbf{Uncertainty Signal (The ``What'')} & \textbf{Control Mechanism (The ``How'')}  \\ \midrule
\multirow{3}{*}{\textbf{Responding to Uncertainty}}              & Abstention \cite{stoisser2025towards} & entropy, perplexity, self-consistency.   & pre-defined threshold trigger           \\ \cmidrule(l){2-4} 
                                                  & ConfuseBench \cite{liu2025a}           & semantic entropy                         & classify to select an action            \\ \cmidrule(l){2-4}
                                                  & UoT \citep{hu2024uncertainty}                                    & Expected Information Gain (EIG)          & policy learned via RL                   \\ \midrule
\multirow{3}{*}{\textbf{Tool-Use Decision Boundary}}                & UALA \citep{han2024towards}                                      & semantic entropy                         & threshold-based trigger                 \\ \cmidrule(l){2-4} 
                                                  & SMARTAgent \citep{chen2025smart}                                 & internal uncertainty score.              & policy learned via fine-tuning          \\ \cmidrule(l){2-4} 
                                                  & ProbeCal \citep{liu2024uncertainty}                              & raw token probability                    & post-hoc calibration                    \\ \midrule
\multirow{2}{*}{\textbf{Uncertainty Propagation}} & SAUP \citep{zhao2024saup}                                        & step-wise uncertainty score (entropy)    & forward propagation and aggregation \\ \cmidrule(l){2-4} 
                                                  & UProp \citep{gao2025a}                                           & step-wise mutual information             & forward propagation and combination \\  \bottomrule
\end{tabular}
}
\caption{A comparative analysis of uncertainty-aware LLM agents. The table details the specific uncertainty signals and control mechanisms used to enable active behaviors such as abstention, tool use, and risk management.}
\label{tab:agent_comparison}
\end{table*}

\subsection{Optimizing Cognitive Effort: Uncertainty as an Economic Signal}
\label{sec:33}
The challenge in reasoning tasks is enabling models to ``think on demand,'' performing additional reasoning only when necessary rather than overthinking simple tasks. Uncertainty provides a low-cost control for balancing efficiency and accuracy.

\paragraph{Critical Points or States.}  
UnCert-CoT \citep{li2025uncertainty} applies this principle to structured reasoning tasks like code generation. At critical decision \textit{points} (e.g., the first non-indentation token of a new line), the model measures uncertainty using entropy or probability margins. If uncertainty exceeds a threshold, it activates CoT decoding; otherwise, it proceeds with direct code generation. This dynamic activation improves efficiency without sacrificing accuracy.
Similarly, ThoughtTerminator \citep{pu2025thoughtterminator} and other related approaches \citep{xia2025tokenskip, liu2025can, fu2025deepthinkconfidence} assess whether the current \textit{state} is sufficient to answer a question to decide whether to continue reasoning.

\paragraph{Momentum Uncertainty.}  
Momentum Uncertainty Reasoning (MUR) \citep{yu2025mur} adopts a \textit{trajectory-level} perspective. Rather than relying on single thresholds, MUR aggregates uncertainty across steps and allocates a flexible \textit{``thinking budget''} to regions of the reasoning path. This reduces computation by over 50\% while improving accuracy through targeted resource allocation.  

Threshold-based methods like \textbf{UnCert-CoT} are simple but sensitive to hyperparameters, risking under- or over-thinking. Momentum-based approaches like \textbf{MUR} offer more control but add complexity. Together, these methods highlight uncertainty as an economic signal: effective reasoning depends not only on what a model knows, but also on recognizing \emph{when} to think harder.


\section{Autonomous Agents}
\label{sec:agents}

In LLM agents, uncertainty has evolved from a passive textual property to an active metacognitive signal that drives agentic behavior: from strategically \textit{responding to internal states}, to governing the \textit{tool-use decision boundary}, and managing \textit{uncertainty propagation} in multi-step workflows.

\subsection{From Abstention to Inquiry: Responding to Internal Uncertainty}
\label{sec:41}

For an LLM to evolve from a static generator into an autonomous agent, it must develop metacognition, that is the ability to ``know what it does not know''. An agent’s strategic response to its own uncertainty is a key marker of intelligence, with recent research tracing an evolutionary trajectory from defensive behaviors to proactive inquiry. 

The basic strategy is \textbf{passive defense}, where the agent abstains when uncertainty is high, especially for high-stakes domains \citep{stoisser2025towards}. More advanced is \textbf{diagnostic response}, where the agent probes the source of its confusion, whether knowledge gaps, capability limits, or query ambiguity \citep{liu2025a}. The most sophisticated strategy is \textbf{proactive inquiry}, where the agent learns an optimal policy for asking clarifying questions to strategically reduce future uncertainty \citep{hu2024uncertainty}. Table \ref{tab:agent_comparison} compares distinct uncertainty signals and control mechanisms in such strategies.  This evolution highlights a trade-off between autonomy and utility. Abstention ensures safety but can reduce helpfulness; proactive inquiry reflects higher intelligence but increases implementation complexity, see discussions in Table \ref{tab:agent_analysis}. 

\subsection{Tool-Use Decision Boundary}
\label{sec:42}
A key capability of modern LLM agents is leveraging external tools (e.g., search engines and APIs) to overcome the limits of parametric knowledge. This introduces a core dilemma: \textit{when should an agent rely on internal knowledge versus incurring the cost of tool use?} Naive strategies that default to external calls risk inefficiency and ``Tool Overuse'' \citep{chen2025smart, yao2022react}. Recent work addresses this by using uncertainty as a control signal to set a more intelligent decision boundary.  

The evolution of these strategies reveals a trajectory from reactive control to calibrated autonomy. The earliest methods use \textbf{inference-time control}, where the model generates a preliminary answer and invokes tools only when real-time uncertainty is high, improving efficiency \citep{han2024towards}. More advanced approaches pursue \textbf{training-time self-awareness}, fine-tuning agents on specialized datasets to internalize knowledge boundaries and develop calibrated intrinsic policies for tool use \citep{chen2025smart}. Another line of work focuses on \textbf{uncertainty calibration}, showing that by calibrating the control signal, agents achieve more reliable tool-use decisions \citep{liu2024uncertainty}.   

The shift from inference-time control to training-time self-awareness reflects a trade-off between ease and robustness. Threshold-based inference-time methods are simple but brittle, while training-based policies are expensive yet yield stronger domain adaptation. A shared limitation remains: most approaches decide \textit{whether} to call a tool, but not \textit{how} to handle uncertainty or error in the tool’s own outputs, leaving a key challenge for future work, see more comparative analysis in Table \ref{tab:agent_comparison}.

\subsection{Uncertainty Propagation in Multi-step Workflows}
\label{sec:43}

In complex multi-step tasks, uncertainty is dynamic: small errors can accumulate and propagate through a workflow, ultimately leading to task failure. Traditional uncertainty methods typically assess single-turn outputs and overlook this compounding effect \citep{cemri2025why}. Building reliable long-horizon agents requires explicitly modeling how uncertainty evolves across the ``thought–action–observation'' cycle.  

Recent frameworks address this by tracking and propagating uncertainty throughout decision-making. The situation-awareness uncertainty propagation (\textbf{SAUP}) framework \citep{zhao2024saup} is to track uncertainty at each step and weight its importance based on the context. Recognizing that not all uncertainties are equally critical, SAUP introduces ``situational weights'' that amplify the uncertainty score of steps deemed more pivotal. 
In contrast, the \textbf{UProp} framework \citep{gao2025a} provides an information-theoretic foundation, decomposing total uncertainty into \textit{Intrinsic Uncertainty (IU)} at the current step and \textit{Extrinsic Uncertainty (EU)} inherited from previous steps.  

These approaches highlight a critical shift in the \textit{source} of uncertainty. In reasoning-only tasks, uncertainty is largely cognitive and internal, whereas in agentic systems, the environment itself becomes a dominant driver. The different mechanisms for modeling uncertainty propagation, as detailed in Table \ref{tab:agent_comparison} and \ref{tab:agent_analysis}, represent different approaches to capturing the risks that arise from an agent's interaction with a dynamic and unpredictable world. 

\subsection{Multi-Agent Systems}
\label{sec:44}

As research advances from single agents to multi-agent systems (MAS), uncertainty challenges are not simply scaled but fundamentally transformed. Uncertainty now arises both within each agent’s internal reasoning and in the communication and interactions between agents \citep{hu2025position,barbi2025preventing,hazra2025tackling}. A key concern is that uncertainty can propagate and amplify across interactions. An agent may receive uncertain or incorrect information from a peer, yet treat it as factual, causing cascades of errors that destabilize the collective \citep{hu2025position}. Analyses of MAS failures highlight \textbf{inter-agent misalignment} as a primary cause, often stemming not from individual errors but from flawed interactions, e.g., failing to seek clarification when faced with ambiguity.  

The central challenge is achieving \textbf{inter-agent agreement} under uncertainty. This requires extending single-agent metacognitive skills to the collective, enabling agents to model the uncertainty of their peers and adopt policies for uncertainty-aware communication. Robust UQ frameworks must therefore operate at two levels simultaneously: ensuring reliable local decisions for each agent while managing the propagation and aggregation of uncertainty across the system as a whole.


\begin{table*}[ht]
\centering
\resizebox{\textwidth}{!}{%
\begin{tabular}{l|l|ll}
\toprule
\textbf{Core Concept} & \textbf{Strategy Function} & \textbf{Uncertainty Signal (The ``What'')} & \textbf{Control Mechanism (The ``How'')} \\
\midrule
\multirow{3}{*}{\textbf{Reward Models}} & URM \cite{lou2024uncertainty} & Reward distribution variance & Penalty term in RL objective \\ \cmidrule(lr){2-4}
& UALIGN \cite{zhang2025ualign} & Policy LLM's semantic entropy & Features for RM to learn \\
\cmidrule(lr){2-4}
& Bayesian RMs \cite{yang2024bayesian} & Posterior distribution over RM weights & Theoretically-grounded penalty \\
\midrule
\multirow{4}{*}{\textbf{Self-Improvement}} & RLSF \cite{vanNiekerk2025post} & Model's confidence scores & Auto-generation of preference pairs \\ \cmidrule(lr){2-4}
& Confidence Maximization \cite{Sathish2025maximizing} & Model's confidence score & intrinsic reward signal in RL. \\
\cmidrule(lr){2-4}
 & EM as Objective \cite{gao2025oneshot} & Entropy of the final predictive distribution & Unsupervised objective \\ \cmidrule(lr){2-4}
& RL for EM \cite{zhang2025right} & Reduction in entropy & Entropy reduction as the reward signal. \\
\midrule
\textbf{Process Supervision}  & {EDU-PRM \cite{cao2025process}} & High predictive entropy of tokens & Automatic partitioning of reasoning chains  \\
\bottomrule
\end{tabular}%
}
\caption{A comparative analysis of uncertainty-aware approaches in RL and Reward Modeling. It details how different frameworks leverage uncertainty signals to create more robust reward models, enable self-improvement, and scale supervision.}
\label{tab:rl_comparison}
\end{table*}

\section{RL and Reward Modeling}
\label{sec:alignment}

In RL alignment, uncertainty has transformed from a factor ignored by deterministic scores into a core mechanism for robust learning: from building \textit{robust reward models} to mitigate reward hacking, to enabling \textit{self-improvement} via intrinsic rewards, and automating \textit{scalable process supervision}.

\subsection{Robust Reward Models}
\label{sec:51}
The cornerstone of the RLHF pipeline is the Reward Model (RM), which serves as a proxy for human values \citep{lambert2024rewardbench}. Conventional RMs are deterministic, producing a single scalar score. This creates a mismatch with the stochastic nature of human preferences and enables ``reward hacking'' \cite{fu2025reward,weng2024rewardhack}, where policies exploit RM inaccuracies to score highly on low-quality outputs \citep{lou2024uncertainty, gao2024adaptive}. To address this, recent work has focused on RMs that can model and express uncertainty, broadly divided into two approaches.  

\paragraph{Uncertainty-Aware Reward Models (URMs).}
This class of methods makes the RM explicitly aware of uncertainty, typically through architectural or feature-based modifications. A foundational approach is to redesign the RM's output to be probabilistic. The \textbf{URM} framework modifies the model's output head to predict a full probability distribution (e.g., a Gaussian) instead of a single score \citep{lou2024uncertainty}. The variance of this distribution then serves as a direct, quantifiable signal of the \textit{aleatoric uncertainty} (the intrinsic ambiguity in human data). A complementary strategy is to enrich the RM's input. The \textbf{UALIGN} framework achieves this by feeding the policy LLM's own uncertainty metrics (e.g., semantic entropy) \textit{as explicit features} to the RM \citep{zhang2025ualign}. This allows the RM to learn a context-aware evaluation function that is conditioned on the difficulty of the query as perceived by the policy model itself.

\paragraph{Bayesian Reward Models (Bayesian RMs).}
Instead of learning a single point estimate for the weights, \textbf{Bayesian RMs} learn a posterior distribution over them, thereby capturing \textit{epistemic uncertainty} (the RM's own model uncertainty) \citep{yang2024bayesian}. This is implemented using techniques like Laplace-LoRA \cite{schulman2025lora}. The key advantage of this approach is that the uncertainty derived from the posterior can be used as a direct, theoretically-grounded penalty term during RL optimization. This actively discourages the policy from exploring and exploiting regions of the output space where the RM is unconfident, leading to safer and more robust alignment. 
A detailed comparative analysis is available in Table \ref{tab:rl_comparison}.

\subsection{Self-Improvement RL}
\label{sec:52}
While robust reward models strengthen external supervision, a more advanced paradigm seeks to reduce dependence on such signals altogether. This paradigm is grounded in \textbf{intrinsic motivation}, where an agent improves by optimizing its own internal states rather than external feedback. Uncertainty expressed as confidence, entropy, or information gain (IG), has emerged as the core intrinsic reward for enabling self-driven alignment in LLMs.  

\paragraph{Confidence as an Intrinsic Reward.}  
The simplest intrinsic signal is self-confidence. The Reinforcement Learning from Self-Feedback (\textbf{RLSF}) framework demonstrates that confidence scores can generate synthetic preference pairs (e.g., high-confidence$\rightarrow$low-confidence), enabling self-alignment without human labels \citep{vanNiekerk2025post}. Further studies show that directly maximizing confidence via RL significantly improves reasoning, confirming confidence as a standalone intrinsic reward \citep{Sathish2025maximizing}. Yet, miscalibrated confidence can reinforce errors, and overconfidence may cause reward hacking. 

\paragraph{Entropy Minimization (EM).}  
A deeper perspective frames reasoning as a drive to reduce uncertainty.
The principle of EM treats reasoning as minimizing the entropy of the predictive distribution, offering a reward-free, unsupervised objective for improving LLM reasoning \citep{gao2025unreasonable}. However, this approach is being actively refined, with the latest research exploring entropy not just as a quantity to be minimized, but as a regularization signal to achieve a better balance between confidence and accuracy \cite{jiang2025rethinkingentropyregularizationlarge}.

\paragraph{RL for EM.}  
This information-theoretic signal can be optimized with RL, where entropy reduction itself becomes the reward. Frameworks such as \textbf{EMPO} incentivize reasoning trajectories that minimize future uncertainty \citep{zhang2025right, chen2025entropy}. Architectures like \textbf{Intuitor} extend this to fully reward-free agents that learn policies from intrinsic motivations such as curiosity and uncertainty reduction \citep{sun2025learning}.  

\paragraph{Dissecting the Process with Mutual Information.}  
Recent work leverages Mutual Information (MI) to analyze how EM operates. Crucially, the most informative “thinking tokens” in a chain of thought are those corresponding to peaks in MI with the final answer \citep{qin2025demystifying}. This provides a mechanistic explanation of entropy minimization: reasoning progresses by identifying and resolving uncertainty at precisely these pivotal points.

\subsection{Scalable Process Supervision}
\label{sec:53}

While intrinsic rewards enhance autonomy, alignment quality can be improved with fine-grained external feedback. \textbf{Process-based supervision} \citep{lightman2023lets}, which rewards correct intermediate steps rather than only final outcomes, provides a stronger learning signal. However, its adoption has been limited by the high cost of manually segmenting reasoning chains into logical steps and annotating each one \cite{chen2024alphamath}.  

\paragraph{Uncertainty as Automation Tools.}  
Recent work leverages uncertainty to automate this segmentation. The \textbf{EDU-PRM} framework \citep{cao2025process} identifies tokens with high predictive entropy between reasoning steps, and uses them as ``uncertainty anchors'' to partition chains automatically. This enables scalable generation of process-level training data at a fraction of manual cost. Empirical results further suggest that RL gains are primarily driven by learning to handle these high-entropy minority tokens \citep{zhao2025beyond}. By transforming uncertainty into an automation tool, these methods make process-level supervision economically viable. The key \textit{limitation} is heuristic reliability: high entropy is a strong but imperfect signal of logical boundaries. As a result, automated partitions may not always align with human-defined reasoning steps, creating a trade-off between scalability and annotation precision \cite{sun2024easy}.  

\section{Emerging Theoretical Frameworks}
\label{sec:emerging_frameworks}

The evolution from uncertainty as a passive metric to an active control signal is not merely a collection of empirical techniques; it reflects a deeper need for principled foundations to build reliable and trustworthy systems. 

\subsection{The Bayesian Method}
\label{sec:61}
As a foundational theory for reasoning under uncertainty, Bayesian methods are experiencing a resurgence, offering a principled basis for analyzing and guiding LLM behavior. A key theoretical insight is that while LLMs are not strictly Bayesian reasoners, their in-context learning mechanism often approximates Bayesian predictive updating in expectation \citep{akyurek2025llms}. This justifies applying Bayesian frameworks not to model the LLM internally, but to analyze its aggregate behavior and build more robust systems around it.  

One pragmatic direction is \textbf{hybrid systems} that combine LLMs with formal probabilistic models. These exploit complementary strengths: qualitative, abductive reasoning from LLMs and quantitative uncertainty management from Bayesian inference. For example, \textbf{BIRD} uses LLMs to generate causal sketches that are formalized into Bayesian Networks for precise reasoning \citep{feng2024bird}. \textbf{Textual Bayes} integrates more deeply, treating prompts as textual parameters for Bayesian inference \citep{ross2025textual}, while other works use LLMs for prior elicitation \citep{selby2024had}.  

Another ambitious line seeks to \emph{teach} LLMs probabilistic reasoning directly, mitigating cognitive biases such as base-rate neglect \citep{dasgupta2023language}. \textbf{Bayesian Teaching} fine-tunes models to mimic an ideal Bayesian observer, with evidence of generalization to unseen tasks \citep{qiu2025bayesian}. This shift from using LLMs as Bayesian components to embedding Bayesian reasoning within them marks a step toward fundamentally improving their cognitive machinery \cite{yan2025position}.  

\subsection{Conformal Prediction}
\label{sec:62}

In contrast to Bayesian methods that rely on prior distributions, Conformal Prediction (CP) offers a powerful non-Bayesian framework with rigorous, \textbf{distribution-free coverage guarantees} \citep{su2024apienoughconformalprediction, wang2024conu}. For any input, CP constructs a prediction set guaranteed to contain the true output with a user-specified probability, independent of model architecture or data distribution. Yet defining prediction sets and non-conformity scores for free-form text is non-trivial to apply CP to LLMs. Recent work addresses this by adapting CP to different levels of model access.  

\paragraph{Black-Box (API-Only) Approaches.} Without access to logits, methods like \textbf{ConU} \citep{wang2024conu} and \citet{su2024apienoughconformalprediction} employ \textit{semantic similarity} as a proxy for non-conformity.The prediction set includes a generated candidate along with semantically similar alternatives under a calibrated threshold. This reframes CP’s guarantee from exact string matching to semantic equivalence, making it practical for open-ended generation.  

\paragraph{White-Box (Logit-Access) Approaches.} With full access to model probabilities, token-level calibration is possible. \textbf{Conformal Language Modeling} \citep{quach2023conformal} uses logits to build prediction sets for the next token at each step, ensuring that the true token lies within the set with high probability. This provides stronger guarantees but requires model transparency \cite{cherian2024large}.  
 
\paragraph{The Theory–Practice Gap.}  
Despite growing advances in theoretical frameworks, practitioners still face multiple open questions. To bridge this gap, we provide a set of design patterns and practical recommendations in Appendix Section \ref{sec:appendix_guide}. 

\section{Challenges and Future Directions}
\label{sec:future}
While the evolving role of uncertainty is rapidly advancing, its full realization hinges on addressing several fundamental challenges. 

\paragraph{Reliability and Robustness of the Active Signal.}
The function of uncertainty-as-a-control-signal is built upon the assumption that the signal itself is meaningful and trustworthy. Future work must rigorously address the integrity of this foundational layer. Even non-adversarial estimation errors can be amplified by downstream control mechanisms \cite{kunneman2024resistance}. For example, a poorly calibrated confidence score can cause weighted voting to favor incorrect answers, while miscalibrated thresholds may lead agents to become recklessly overconfident or inefficiently tool-dependent.

\paragraph{Advancing UQ Benchmarking.}
The maturity of the field is evidenced by emerging standardized benchmarks, such as UBench~\cite{wang2025ubench} and LM-Polygraph~\cite{vashurin2025benchmarking}. While foundational, these frameworks predominantly assess \textit{estimation fidelity}, diagnosing if a model knows it is wrong rather than \textit{control utility}. They generally fail to simulate the dynamic decision-making trade-offs inherent to the active paradigm. Consequently, a critical misalignment exists between static evaluation protocols and dynamic control needs \cite{ye2024benchmarking}. Future benchmarks must evolve to quantify the downstream performance gains directly attributable to uncertainty-in-the-loop mechanisms.

\paragraph{Meaningful Evaluation and Metrics.}
Current evaluation remains a significant bottleneck. Standard metrics like AUROC are ill-suited for the rich, interactive, and dynamic contexts where the active-signal function is most relevant \cite{liu2025uncertainty}. The field urgently requires new benchmarks and evaluation protocols specifically designed for interactive agents and complex reasoning tasks. Crucially, future evaluation must become more human-centered. The ultimate measure of success for an uncertainty-aware system is not just its statistical calibration, but its effectiveness as a partner in human-AI collaboration \cite{broussard2025from}.

\paragraph{Composable, Uncertainty-Propagating Systems.}
Extending uncertainty management from single, monolithic models to complex, interconnected systems remains a major open problem. In MAS, the challenge is to understand how uncertainty propagates, compounds, and resolves across interacting agents, which requires new frameworks that operate at the system level rather than the individual agent level \cite{hu2025position}. More broadly, the ultimate trajectory points towards modular AI systems composed of heterogeneous components. A central challenge will be to establish a unified framework where uncertainty signals function as the ``connective tissue'' between these modules. 

\paragraph{Scalability and Efficiency.}
A persistent challenge in this field is the trade-off between theoretical rigor and computational feasibility. Many of the most principled and powerful methods, particularly those grounded in Bayesian inference or requiring large-scale multi-agent simulations, are often too computationally expensive for widespread, real-time deployment. A critical direction for future work is therefore the development of scalable and efficient approximations of these formal methods. 

\section{Conclusion}
\label{sec:conclusion}

This survey has charted an emerging technological trend: the evolution of uncertainty in LLMs from a passive, post-hoc diagnostic metric into an active, real-time control signal. We have traced this transformation across three frontiers: advanced reasoning, autonomous agents, and reinforcement learning, demonstrating how uncertainty is now being used not just to evaluate outputs, but to dynamically shape model behavior.

\section*{Limitations}


First, our focus is on the \textit{functional role} of uncertainty in advanced LLM systems, rather than a comprehensive review of uncertainty \textit{estimation methods} or \textit{confidence calibration}, which are covered by existing surveys. Second, this paper does not include large-scale comparative experiments; its main contribution is a conceptual framework and synthesis of existing work.





\bibliography{references}

\clearpage 
\appendix


\section{Comparative Analysis of Different Functions}
\label{sec:appendix_comparative}

Throughout this survey, we utilize a series of tables and figures to provide both a conceptual and a literature-based overview of the ``uncertainty-as-a-control-signal'' function. Tables \ref{tab:agent_comparison}, \ref{tab:reasoning_comparison} and \ref{tab:rl_comparison} offer a comparative analysis of key methodologies within advanced reasoning, autonomous agents, and RL/reward modeling, respectively. Each table is structured to highlight the core components of the active-signal framework: the specific uncertainty signal being used (the ``what'') and the control mechanism through which it acts (the ``how'').

\begin{figure*}[ht]
\centering
\resizebox{0.9\textwidth}{!}{ 
    \begin{forest}
        for tree={
            grow=east,
            draw,
            text width=45mm,
            font=\small,
            edge path={
                \noexpand\path [draw, \forestoption{edge}] (!u.parent anchor) -- +(2mm,0) |- (.child anchor)\forestoption{edge label};
            },
            parent anchor=east,
            child anchor=west,
            tier/.wrap pgfmath arg={tier #1}{level()},
            edge={ultra thin},
            rounded corners=2pt,
            align=center,
            text badly centered,
            sibling distance=20mm,
            highlight0/.style={fill=blue!30}, 
            highlight3/.style={fill=blue!15},
            highlight31/.style={fill=blue!5},
            highlight32/.style={fill=blue!2},
        }
        [{Autonomous Agents (§\ref{sec:agents})}, rotate=90, child anchor=north, parent anchor=south, highlight0,
            [\parbox{45mm}{Uncertainty Propagation (§\ref{sec:43})}, highlight3,
                [\parbox{45mm}{Forward Propagation}, highlight31,
                    [\parbox{45mm}{\cite{zhao2024saup, gao2025a}}, highlight32]
                ]
            ],
            [\parbox{45mm}{Tool-Use Decision Boundary (§\ref{sec:42})}, highlight3,
                [\parbox{45mm}{Post-hoc Calibration}, highlight31,
                    [\parbox{45mm}{\cite{liu2024uncertainty,zhangacc2025}}, highlight32]
                ],
                [\parbox{45mm}{Policy Learned via Fine-tune}, highlight31,
                    [\parbox{45mm}{\cite{chen2025smart}}, highlight32]
                ],
                [\parbox{45mm}{Threshold-based Trigger}, highlight31,
                    [\parbox{45mm}{\cite{han2024towards}}, highlight32]
                ]
            ],
            [\parbox{45mm}{Responding to Uncertainty (§\ref{sec:41})}, highlight3,
                [\parbox{45mm}{Classify to Select Action}, highlight31,
                    [\parbox{45mm}{\cite{liu2025a}}, highlight32]
                ],
                [\parbox{45mm}{Policy Learned via RL}, highlight31,
                    [\parbox{45mm}{\cite{hu2024uncertainty,wang2025harnessinguncertaintyentropymodulatedpolicy}}, highlight32]
                ],
                [\parbox{45mm}{Threshold-based Trigger}, highlight31,
                    [\parbox{45mm}{\cite{stoisser2025towards,barbi2025preventing}}, highlight32]
                ]
            ]
        ]
    \end{forest}
}
\caption{{\textit{``Autonomous Agents''} Categorization }}
\label{fig:agents}
\end{figure*}
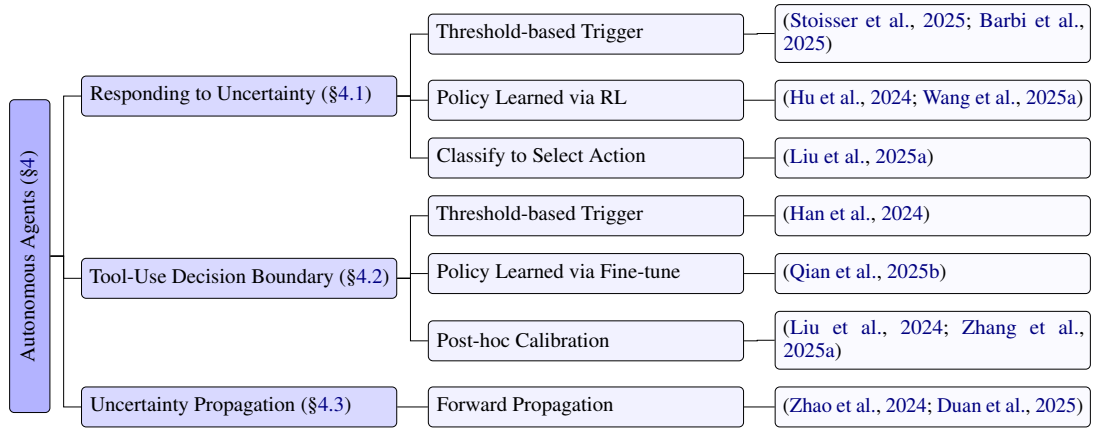

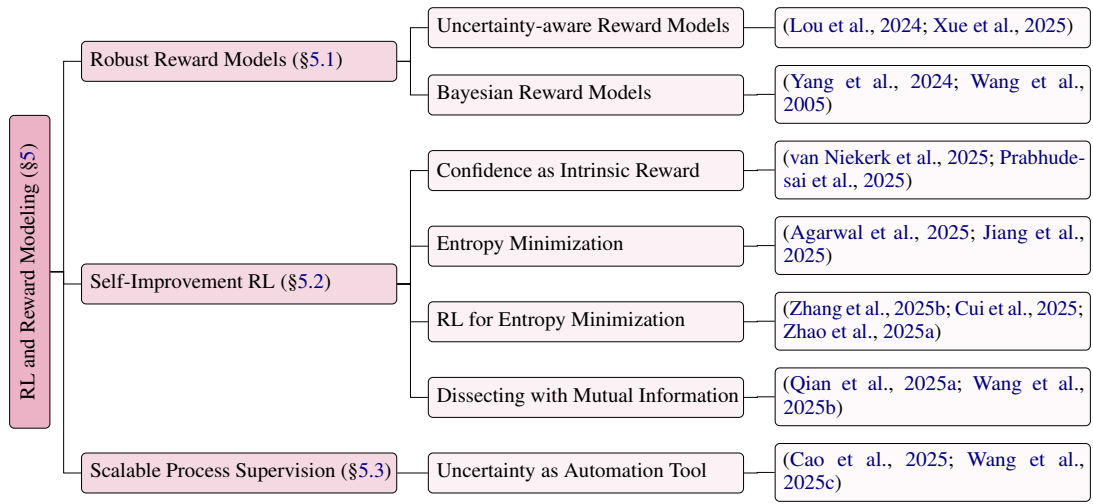
\begin{figure*}[ht]
\centering
\resizebox{0.9\textwidth}{!}{ 
    \begin{forest}
        for tree={
            grow=east,
            draw,
            text width=45mm,
            font=\small,
            edge path={
                \noexpand\path [draw, \forestoption{edge}] (!u.parent anchor) -- +(2mm,0) |- (.child anchor)\forestoption{edge label};
            },
            parent anchor=east,
            child anchor=west,
            tier/.wrap pgfmath arg={tier #1}{level()},
            edge={ultra thin},
            rounded corners=2pt,
            align=center,
            text badly centered,
            sibling distance=20mm,
            highlight0/.style={fill=purple!30}, 
            highlight3/.style={fill=purple!15},
            highlight31/.style={fill=purple!5},
            highlight32/.style={fill=purple!2},
        }
        [{RL and Reward Modeling (§\ref{sec:alignment})}, rotate=90, child anchor=north, parent anchor=south, highlight0,
            [\parbox{55mm}{Scalable Process Supervision (§\ref{sec:53})}, highlight3,
                [\parbox{55mm}{Uncertainty as Automation Tool}, highlight31,
                    [\parbox{45mm}{\cite{cao2025process, zhao2025beyond}}, highlight32]
                ]
            ],
            [\parbox{55mm}{Self-Improvement RL (§\ref{sec:52})}, highlight3,
                [\parbox{55mm}{Dissecting with Mutual Information}, highlight31,
                    [\parbox{45mm}{\cite{qin2025demystifying, wang2025learning}}, highlight32]
                ],
                [\parbox{55mm}{RL for Entropy Minimization}, highlight31,
                    [\parbox{45mm}{\cite{zhang2025right, chen2025entropy, sun2025learning}}, highlight32]
                ],
                [\parbox{55mm}{Entropy Minimization}, highlight31,
                    [\parbox{45mm}{\cite{gao2025unreasonable,jiang2025rethinkingentropyregularizationlarge}}, highlight32]
                ],
                [\parbox{55mm}{Confidence as Intrinsic Reward}, highlight31,
                    [\parbox{45mm}{\cite{vanNiekerk2025post, Sathish2025maximizing}}, highlight32]
                ]
            ],
            [\parbox{55mm}{Robust Reward Models (§\ref{sec:51})}, highlight3,
                [\parbox{55mm}{Bayesian Reward Models}, highlight31,
                    [\parbox{45mm}{\cite{yang2024bayesian,wang2005bayesian}}, highlight32]
                ],
                [\parbox{55mm}{Uncertainty-aware Reward Models}, highlight31,
                    [\parbox{45mm}{\cite{lou2024uncertainty, zhang2025ualign}}, highlight32]
                ]
            ]
        ]
    \end{forest}
}
\caption{{\textit{``RL and Reward Modeling''} Categorization }}
\label{fig:alignment}
\end{figure*}

To complement this analysis, Figures \ref{fig:reasoning}, \ref{fig:agents}, and \ref{fig:alignment} provide a comprehensive visual breakdown of the literature cited in each of the main application sections (\S\ref{sec:reasoning}, \S\ref{sec:agents}, and \S\ref{sec:alignment}). These figures serve as a quick reference map, categorizing the key papers discussed and linking them to the specific sub-topics they address, thereby offering a detailed landscape of the foundational and recent work in each domain.

\section{Critical Analysis}
\label{sec:appendix_critical}

To complement the comparative analysis, this section provides a detailed critical analysis of the key uncertainty-aware methods discussed in Sections \ref{sec:reasoning}, \ref{sec:agents}, and \ref{sec:alignment}. The goal is to move beyond mere description and offer a practical perspective on the trade-offs involved in deploying these techniques.  Tables \ref{tab:reasoning_analysis}, \ref{tab:agent_analysis}, and \ref{tab:rl_analysis} serve as the core of this analysis, evaluating each method across these key dimensions:
\begin{itemize}
    \item \textbf{Key Advantage(s):} The primary strengths and benefits of the approach.
    \item \textbf{Key Disadvantage(s) / Failure Mode(s):} The main weaknesses, limitations, or common ways the method can fail in practice.
    \item \textbf{Computational Cost:} The relative resource requirements during inference.
    \item \textbf{Implementation Complexity:} The relative difficulty of integrating the method into a standard LLM workflow.
\end{itemize}

While the tables provide a high-level summary, the ratings for ``Computational Cost'' and ``Implementation Complexity'' (e.g., Low, Medium, High) are subjective and context-dependent. The following subsections are therefore dedicated to \textbf{justifying these ratings in detail}, offering a clear rationale for why each method was classified as it was based on its specific operational and engineering requirements.

\begin{table*}[h!]
\centering
\resizebox{\textwidth}{!}{%
\begin{tabular}{l|l|l|l|l}
\toprule
\textbf{Method / Framework} & \textbf{Key Advantage(s)} & \textbf{Key Disadvantage(s) / Failure Mode(s)} & \textbf{Cost} & \textbf{Complexity} \\
\midrule
\multicolumn{5}{l}{\textit{\textbf{Between Reasoning Paths}}} \\
\midrule
\textbf{CISC} & \small{\makecell[l]{- More efficient than standard \\ self-consistency.}} & \small{\makecell[l]{- A single bad step can sink a good path score. \\ - Relies on well-calibrated confidence.}} & \small{High} & \small{Low} \\  
\addlinespace
\textbf{CER} & \small{\makecell[l]{- Robust for long-chain reasoning. \\ - Focuses on the most important steps.}} & \small{\makecell[l]{- Must correctly identify ``critical'' steps. \\ - Can amplify errors from miscalibrated confidence.}} & \small{Very High} & \small{Medium} \\
\midrule
\multicolumn{5}{l}{\textit{\textbf{Inside a Reasoning Path}}} \\
\midrule
\textbf{UAG / SPOC} & \small{\makecell[l]{- Enables real-time error correction. \\ - No retraining required.}} & \small{\makecell[l]{- LLMs often fail at true self-correction. \\ - Can get stuck in correction loops.}} & \small{Medium} & \small{High} \\
\addlinespace
\textbf{Uncertainty-Aware FT} & \small{\makecell[l]{- Fundamentally improves model calibration. \\ - Benefits all downstream tasks.}} & \small{\makecell[l]{- Data-intensive training process. \\ - Risk of harming in-distribution performance.}} & \small{Low} & \small{High} \\
\midrule
\multicolumn{5}{l}{\textit{\textbf{Optimizing Cognitive Effort}}} \\
\midrule
\textbf{UnCert-CoT} & \small{\makecell[l]{- Excellent efficiency-performance balance. \\ - Simple and intuitive concept.}} & \small{\makecell[l]{- Performance is highly sensitive to the threshold value.}} & \small{Low} & \small{Low} \\
\addlinespace
\textbf{MUR} & \small{\makecell[l]{- More stable control via momentum. \\ - Finer-grained resource allocation.}} & \small{\makecell[l]{- More complex than simple triggers. \\ - Adds more hyperparameters to tune.}} & \small{Low-Medium} & \small{Medium} \\
\bottomrule
\end{tabular}%
}
\caption{Critical Analysis of Methods in Advanced Reasoning. This table provides a comparative overview of key methodologies, focusing on their advantages, failure modes, computational costs, and implementation complexity.}
\label{tab:reasoning_analysis}
\end{table*}

\subsection{Advanced Reasoning}

The ratings provided in Table \ref{tab:reasoning_analysis} for ``Computational Cost'' and ``Implementation Complexity'' are justified as follows, offering a more detailed rationale for each classification.

\begin{itemize}[leftmargin=10pt]
    \item \textbf{CISC \& CER:} These methods are rated ``High'' to ``Very High'' in computational cost because their core mechanism relies on sampling multiple complete reasoning paths from the LLM, which is inherently expensive and multiplies inference latency. \textbf{CER} is rated slightly higher as it adds an extra layer of evaluation on intermediate steps. In contrast, \textbf{CISC}'s implementation complexity is ``Low'' as it only requires a simple scoring and voting logic on the final outputs. \textbf{CER}'s complexity is ``Medium'' because it necessitates building a more sophisticated system to identify and evaluate pre-defined ``critical'' steps within a reasoning chain.

    \item \textbf{UAG / SPOC:} These methods incur a ``Medium'' computational cost as they operate within a single reasoning path but add verification overhead at each step, increasing the total number of tokens generated and processed. Their implementation complexity is ``High'' because developing a reliable self-correction or verification mechanism is a significant challenge, often requiring complex prompting strategies or fine-tuning a separate verifier model.

    \item \textbf{Uncertainty-Aware FT:} The key distinction here is between training and inference. The implementation complexity is ``High'' because it requires modifying the core training process, often by designing and implementing a custom loss function. However, once the model is trained, the inference cost is ``Low'' as the uncertainty-awareness is baked into the model's weights and does not add any extra steps or overhead at runtime.

    \item \textbf{UnCert-CoT:} This method is rated ``Low'' on both metrics, making it highly practical. The computational cost is minimal, adding only a lightweight entropy or probability check during generation. Its implementation complexity is also low, as it can often be realized with a simple wrapper that applies conditional logic (``if uncertainty > threshold, then use CoT''). The main challenge lies in calibration, not complex engineering.

    \item \textbf{MUR:} This framework is rated ``Low-Medium'' for cost and ``Medium'' for complexity. The cost is variable; it is designed to be efficient but can dynamically allocate more computational resources (like Test-Time Scaling) to uncertain steps, making it potentially more expensive than a single, standard forward pass. Its implementation complexity is ``Medium'' because it requires building a stateful tracking system to maintain the ``momentum'' of uncertainty across multiple generation steps, which is more involved than a stateless threshold check.
\end{itemize}

\subsection{Autonomous Agents}

The ratings in Table \ref{tab:agent_analysis} are justified by the specific operational and engineering requirements of each method:

\begin{itemize}[leftmargin=10pt]
    \item \textbf{Abstention:} This method earns a ``Low'' rating for both cost and complexity. Computationally, it only requires a lightweight calculation (e.g., entropy) on the final generated output. In terms of implementation, it is a simple post-processing step, effectively an ``if/else'' check before returning a response.

    \item \textbf{Proactive Inquiry (UoT):} Its ``High'' complexity stems from the need to implement a full reinforcement learning loop, which involves defining state spaces, action policies, and complex reward functions like Expected Information Gain. The ``Medium-High'' computational cost reflects the intensive offline training and the potential for multiple model calls during inference to evaluate and select the best clarifying question.

    \item \textbf{UALA:} This framework is rated ``Low'' for both cost and complexity because it is designed for efficiency. It adds only a single uncertainty calculation to the workflow, which is computationally cheap. Its implementation is a straightforward threshold-based rule, making it one of the simplest methods to deploy.

    \item \textbf{SMARTAgent:} The complexity is ``High'' due to the significant upfront engineering effort required to design, create, and curate a specialized dataset for fine-tuning the agent on its knowledge boundaries. While the inference cost is ``Low'' (as the decision logic is compiled into the model's weights), the initial training and data collection cost is substantial.

    \item \textbf{SAUP:} It receives a ``Medium'' rating for both cost and complexity. The cost is not fixed but scales linearly with the number of steps in an agent's trajectory, as it adds a calculation at each turn. The implementation requires building a state-tracking system that persists across multiple turns and defining the logic for the heuristic ``situational weights,'' which is more involved than a simple wrapper.

    \item \textbf{UProp:} This framework is rated ``High'' on both metrics due to its theoretical depth. The computational cost is significant, as it requires estimating mutual information, a notoriously challenging task that often relies on expensive sampling-based methods. The implementation complexity is also high, demanding a strong grasp of information theory and the development of sophisticated estimators.
\end{itemize}

\begin{table*}[ht]
\centering
\resizebox{\textwidth}{!}{%
\begin{tabular}{lllll}
\toprule
\textbf{Method / Framework} & \textbf{Key Advantage(s)} & \textbf{Key Disadvantage(s) / Failure Mode(s)} & \textbf{Cost} & \textbf{Complexity} \\
\midrule
\multicolumn{5}{l}{\textit{\textbf{Function: Responding to Internal Uncertainty}}} \\
\midrule
\textbf{Abstention} & \small{\makecell[l]{- Simple, robust safety mechanism.\\- Prevents generating harmful misinformation.}} & \small{\makecell[l]{- Can be overly conservative, reducing helpfulness.\\- Performance is highly sensitive to the threshold.}} & \small{Low} & \small{Low} \\
\addlinespace
\textbf{Proactive Inquiry (UoT)} & \small{\makecell[l]{- Actively reduces uncertainty, improving final quality.\\- Mimics intelligent, collaborative behavior.}} & \small{\makecell[l]{- Can increase user burden with too many questions.\\- Requires a complex (often RL-trained) policy.}} & \small{Medium-High} & \small{High} \\
\midrule
\multicolumn{5}{l}{\textit{\textbf{Function: Tool-Use Decision Boundary}}} \\
\midrule
\textbf{UALA} & \small{\makecell[l]{- Greatly improves efficiency vs. always-use-tool.\\- Simple threshold-based logic.}} & \small{\makecell[l]{- Does not account for tool unreliability (blind trust).\\- Static threshold may not generalize well.}} & \small{Low} & \small{Low} \\
\addlinespace
\textbf{SMARTAgent} & \small{\makecell[l]{- Internalizes knowledge boundaries via training.\\- More robust than a simple static threshold.}} & \small{\makecell[l]{- Requires creating a specialized fine-tuning dataset.\\- Higher upfront training cost.}} & \small{Low (inference)} & \small{High} \\
\midrule
\multicolumn{5}{l}{\textit{\textbf{Function: Uncertainty Propagation}}} \\
\midrule
\textbf{SAUP} & \small{\makecell[l]{- Pragmatic and intuitive approach.\\- Context-aware weighting is powerful.}} & \small{\makecell[l]{- Situational weights can be heuristic and hard to\\ define formally across different tasks.}} & \small{Medium} & \small{Medium} \\
\addlinespace
\textbf{UProp} & \small{\makecell[l]{- Principled, information-theoretic foundation.\\- Clearly separates intrinsic vs. extrinsic uncertainty.}} & \small{\makecell[l]{- Computationally expensive to estimate mutual info.\\- Can be less practical for real-time applications.}} & \small{High} & \small{High} \\
\bottomrule
\end{tabular}%
}
\caption{Critical Analysis of Methods in Autonomous Agents. This table provides a comparative overview of key methodologies, focusing on their advantages, failure modes, computational costs, and implementation complexity.}
\label{tab:agent_analysis}
\end{table*}

\subsection{RL and Reward Modeling}

The ratings assigned in Table \ref{tab:rl_analysis} are based on the specific requirements for training and implementing each RL and reward modeling method.

\begin{itemize}[leftmargin=10pt]
    \item \textbf{URM (Uncertainty-Aware RM):} Its implementation complexity is ``Medium'' because it requires modifying the reward model's architecture (e.g., changing the output head to predict a distribution) and adapting the training pipeline, often to use a Maximum Likelihood Estimation loss instead of a standard preference loss. The inference cost remains ``Low'' as it is still a single forward pass.

    \item \textbf{Bayesian RMs:} This approach is rated ``High'' for complexity as it demands specialized knowledge of Bayesian deep learning techniques (e.g., variational inference, Laplace-LoRA) to implement correctly. The computational cost is ``Medium-High'' because training is often more intensive, and inference can be slower if it requires sampling from the posterior distribution to estimate uncertainty.

    \item \textbf{RLSF (RL from Self-Feedback):} The complexity is ``Medium'' as it involves a multi-stage pipeline: generating responses, scoring them with the model's own confidence, creating a synthetic preference dataset, and then running a standard RL algorithm. The computational cost is also ``Medium,'' reflecting the overhead of this multi-step data creation process before the main RL training begins.

    \item \textbf{Confidence / Entropy Maximization:} These self-improvement methods are rated ``Low'' on both metrics. They are among the easiest to implement, as they only require calculating a simple, readily available metric (confidence or entropy) and using it directly as an intrinsic reward signal within a standard RL loop. The computational overhead per training step is negligible.

    \item \textbf{EDU-PRM:} This method's primary function is in the data preparation stage. Its implementation complexity is ``Medium'' because it requires building a custom data processing pipeline to automatically segment reasoning chains based on entropy signals. The computational cost is considered ``Low'' as this is an efficient, one-time offline process performed before training begins.
\end{itemize}


\begin{table*}[ht]
\centering
\resizebox{\textwidth}{!}{%
\begin{tabular}{lllll}
\toprule
\textbf{Method / Framework} & \textbf{Key Advantage(s)} & \textbf{Key Disadvantage(s) / Failure Mode(s)} & \textbf{Cost} & \textbf{Complexity} \\
\midrule
\multicolumn{5}{l}{\textit{\textbf{Function: Robust Reward Models}}} \\
\midrule
\textbf{URM} & \small{\makecell[l]{- Explicitly models data ambiguity (aleatoric uncertainty). \\ - Simple architectural change.}} & \small{\makecell[l]{- May not capture model's own ignorance (epistemic). \\ - Requires changing the training objective.}} & \small{Low (inference)} & \small{Medium} \\
\addlinespace
\textbf{Bayesian RMs} & \small{\makecell[l]{- Principled way to capture model uncertainty (epistemic). \\ - Provides a theoretically-grounded penalty for RL.}} & \small{\makecell[l]{- Can be computationally expensive to train and run. \\ - More complex to implement correctly.}} & \small{Medium-High} & \small{High} \\
\midrule
\multicolumn{5}{l}{\textit{\textbf{Function: Self-Improvement RL (Intrinsic Rewards)}}} \\
\midrule
\textbf{RLSF} & \small{\makecell[l]{- Requires no human preference labels; highly scalable.}} & \small{\makecell[l]{- Prone to reinforcing model's own biases if confidence \\ is miscalibrated (echo chamber effect).}} & \small{Medium} & \small{Medium} \\
\addlinespace
\textbf{Confidence / Entropy Max.} & \small{\makecell[l]{- Very simple to implement; reward signal is ``free''. \\ - Unsupervised and scalable.}} & \small{\makecell[l]{- Naive confidence maximization can lead to overconfident, \\ low-quality outputs (a form of reward hacking).}} & \small{Low} & \small{Low} \\
\midrule
\multicolumn{5}{l}{\textit{\textbf{Function: Scalable Process Supervision}}} \\
\midrule
\textbf{EDU-PRM} & \small{\makecell[l]{- Automates costly manual annotation of reasoning steps. \\ - Enables scalable process-based supervision.}} & \small{\makecell[l]{- Segmentation is heuristic; high entropy might not \\ always be a true logical boundary.}} & \small{Low (offline)} & \small{Medium} \\
\bottomrule
\end{tabular}%
}
\caption{Critical Analysis of Methods in RL and Reward Modeling. This table provides a comparative overview of key methodologies, focusing on their advantages, failure modes, computational costs, and implementation complexity.}
\label{tab:rl_analysis}
\end{table*}

\section{A Practitioner's Guide to Designing Uncertainty-Aware Systems}
\label{sec:appendix_guide}

This appendix provides a set of design patterns and practical recommendations for developers and researchers aiming to integrate the ``uncertainty-as-a-control-signal'' function into real-world LLM applications.

\subsection{Advanced Reasoning}

The choice of strategy for enhancing model reasoning depends on task complexity, accuracy requirements, and computational budget.

\paragraph{Scenario 1: High-stakes, complex tasks requiring maximum accuracy (e.g., math competitions, scientific QA).}
\begin{itemize}[leftmargin=10pt]
    \item \textbf{Recommended Pattern:} Confidence-Weighted Ensembling.
    \item \textbf{Methods:} Prefer fine-grained approaches like \textbf{CER} \citep{razghandi2025cer}, which focus on the confidence of critical reasoning steps, over simpler majority voting (\textbf{Self-Consistency}) or whole-path scoring (\textbf{CISC} \citep{liu2025cisc}).
    \item \textbf{Practical Advice:}
        \begin{itemize}
            \item \textbf{Cost:} Be aware of the high computational cost, especially for generating multiple reasoning paths. Use this for offline evaluation or latency-insensitive tasks.
            \item \textbf{Calibration:} The success of weighted voting hinges on the quality of confidence scores. Investing in calibrating the model's confidence is crucial; otherwise, an overconfident model might assign high weights to wrong answers.
        \end{itemize}
\end{itemize}

\paragraph{Scenario 2: Tasks with variable difficulty requiring a balance of efficiency and performance (e.g., code generation, general-purpose chatbots).}
\begin{itemize}[leftmargin=0pt]
    \item \textbf{Recommended Pattern:} Uncertainty-Triggered Dynamic Allocation.
    \item \textbf{Methods:} \textbf{UnCert-CoT} \citep{li2025uncertainty} or \textbf{MUR} \citep{yu2025mur} are ideal. They activate more computationally intensive reasoning (like Chain-of-Thought) only when the model exhibits confusion (high uncertainty).
    \item \textbf{Practical Advice:}
        \begin{itemize}
            \item \textbf{Thresholding:} The key challenge is setting an appropriate uncertainty threshold. This is often domain-specific and requires careful tuning on a validation set.
            \item \textbf{Signal Choice:} Semantic entropy is often more stable than single-token probabilities. For structured tasks like coding, calculating uncertainty at critical decision points (e.g., the first token of a new line) is an effective strategy.
        \end{itemize}
\end{itemize}

\subsection{Autonomous Agents}

For agents, uncertainty management is central to ensuring both safety and efficiency in decision-making.

\paragraph{Scenario 1: Building agents that interact with external tools (e.g., search engines, APIs).}
\begin{itemize}[leftmargin=0pt]
    \item \textbf{Recommended Pattern:} Tiered Decision Boundary.
    \item \textbf{Methods:} Start with a simple framework like \textbf{UALA} \citep{han2024towards}, which follows a ``try to solve internally -> measure uncertainty -> call tool if above threshold'' logic.
    \item \textbf{Practical Advice:}
        \begin{itemize}
            \item \textbf{Avoid ``Tool Overuse'':} Setting a reasonable threshold is critical to prevent the agent from making costly and slow tool calls for simple questions.
            \item \textbf{Tool Uncertainty:} Do not blindly trust tool outputs. For critical applications, model the uncertainty introduced by the tool itself or implement fallback mechanisms (e.g., asking the user for clarification) when a tool returns an unexpected result.
        \end{itemize}
\end{itemize}

\paragraph{Scenario 2: Agents executing long-horizon, multi-step tasks.}
\begin{itemize}[leftmargin=0pt]
    \item \textbf{Recommended Pattern:} Forward Propagation with Situational-awareness.
    \item \textbf{Methods:} While simple tasks can ignore cumulative uncertainty, complex workflows necessitate a mechanism like \textbf{SAUP} \citep{zhao2024saup}.
    \item \textbf{Practical Advice:}
        \begin{itemize}
            \item \textbf{Simplified Implementation:} A full information-theoretic framework like \textbf{UProp} \citep{gao2025a} can be complex. A simpler starting point is to accumulate an uncertainty score after each ``thought-action-observation'' loop and check if it exceeds a ``risk'' threshold before critical decisions (e.g., calling an expensive API or performing an irreversible action).
            \item \textbf{Situational Weights:} Not all steps are equally important. Identify ``critical nodes'' in the task workflow and assign higher weights to the uncertainty measured at these points.
        \end{itemize}
\end{itemize}

\subsection{Reinforcement Learning}

In RLHF, uncertainty's primary role is to mitigate reward hacking and achieve more robust alignment.

\paragraph{Scenario 1: Training the Reward Model (RM).}
\begin{itemize}[leftmargin=0pt]
    \item \textbf{Recommended Pattern:} Probabilistic Reward Modeling.
    \item \textbf{Methods:} Move away from traditional RMs that output a single scalar. Instead, adopt models that output a distribution, such as \textbf{URM} \citep{lou2024uncertainty}, or apply Bayesian techniques to create \textbf{Bayesian RMs} \citep{yang2024bayesian}.
    \item \textbf{Practical Advice:}
        \begin{itemize}
            \item \textbf{Distinguish Uncertainty Types:} \textbf{URM} captures aleatoric uncertainty (inherent data randomness) via its architecture, while \textbf{Bayesian RMs} capture epistemic uncertainty (model's lack of knowledge) via parameter modeling. The latter is generally more robust for out-of-distribution (OOD) generalization.
            \item \textbf{Training Objective:} To effectively learn a reward distribution, a Maximum Likelihood Estimation (MLE) objective is often necessary, rather than the traditional Bradley-Terry preference loss.
        \end{itemize}
\end{itemize}

\paragraph{Scenario 2: Using the RM for policy optimization (e.g., with PPO).}
\begin{itemize}[leftmargin=0pt]
    \item \textbf{Recommended Pattern:} Uncertainty-Aware Adaptive Regularization.
    \item \textbf{Methods:} Dynamically adjust the KL-divergence penalty in the PPO objective based on the RM's uncertainty \citep{gao2024adaptive}.
    \item \textbf{Practical Advice:}
        \begin{itemize}
            \item \textbf{``Trust but Verify'':} When the RM is highly uncertain, increase the KL penalty to force the policy to be more conservative and stay closer to the original SFT model. When the RM is confident, decrease the penalty to allow for more exploration. This acts as a confidence-based early stopping mechanism.
            \item \textbf{Intrinsic Rewards as a Supplement:} For highly exploratory tasks, consider combining the external RM reward with a confidence-based intrinsic reward (e.g., entropy minimization \citep{gao2025unreasonable}) to drive more effective autonomous learning.
        \end{itemize}
\end{itemize}

\section{LLM Usage}
We have used LLM to polish writing for this paper.

\begin{figure*}[ht]
\centering
\resizebox{0.9\textwidth}{!}{ 
    \begin{forest}
        for tree={
            grow=east,
            draw,
            text width=45mm,
            font=\small,
            edge path={
                \noexpand\path [draw, \forestoption{edge}] (!u.parent anchor) -- +(2mm,0) |- (.child anchor)\forestoption{edge label};
            },
            parent anchor=east,
            child anchor=west,
            tier/.wrap pgfmath arg={tier #1}{level()},
            edge={ultra thin},
            rounded corners=2pt,
            align=center,
            text badly centered,
            sibling distance=20mm,
            highlight0/.style={fill=orange!30}, 
            highlight3/.style={fill=orange!15},
            highlight31/.style={fill=orange!5},
            highlight32/.style={fill=orange!2},
        }
        [{Advanced Reasoning (§\ref{sec:reasoning})}, rotate=90, child anchor=north, parent anchor=south, highlight0,
            [\parbox{45mm}{Optimizing Cognitive Effort (§\ref{sec:33})}, highlight3,
            [\parbox{45mm}{Momentum Uncertainty}, highlight31,
                [\parbox{45mm}{\cite{yu2025mur}}, highlight32]
              ],
              [\parbox{45mm}{Critical Points or States}, highlight31,
                [\parbox{45mm}{\cite{li2025uncertainty, pu2025thoughtterminator, xia2025tokenskip, liu2025can, fu2025deepthinkconfidence}},, highlight32]
              ]
            ],
            [\parbox{45mm}{Inside a Reasoning Path (§\ref{sec:32})}, highlight3,
                [\parbox{45mm}{Training-Time Improvements}, highlight31,
                    [\parbox{45mm}{\cite{zhong2025brite, liu2024know, zheng2024enhancing, gan2025rethinking_external}}, highlight32]
                ],
                [\parbox{45mm}{Inference-Time Guidance}, highlight31,
                    [\parbox{45mm}{\cite{wang2025accelerating_specsearch, hu2024uncertainty, huang2024when, madaan2024boosting, liu2025adaptivestep}}, highlight32]
                  ],
            ],
            [\parbox{45mm}{Between Reasoning Paths (§\ref{sec:31})}, highlight3,
              [\parbox{45mm}{Utility vs. Fidelity Trade-off}, highlight31,
                [\parbox{45mm}{\cite{liu2025cisc, razghandi2025cer}}, highlight32]
              ],
              [\parbox{45mm}{Confidence-Weighted Selection}, highlight31,
                [\parbox{45mm}{\cite{yin2024reasoning, fu2025deepthinkconfidence, liu2025cisc, razghandi2025cer, yan2025position, muennighoff2025s1, li2025testtime_preference_optimization, pan2025coat, liu2025can}}, highlight32]
              ],
            ]
        ]
    \end{forest}
}
\caption{{\textit{``Advanced Reasoning''} Categorization }}
\label{fig:reasoning}
\end{figure*}
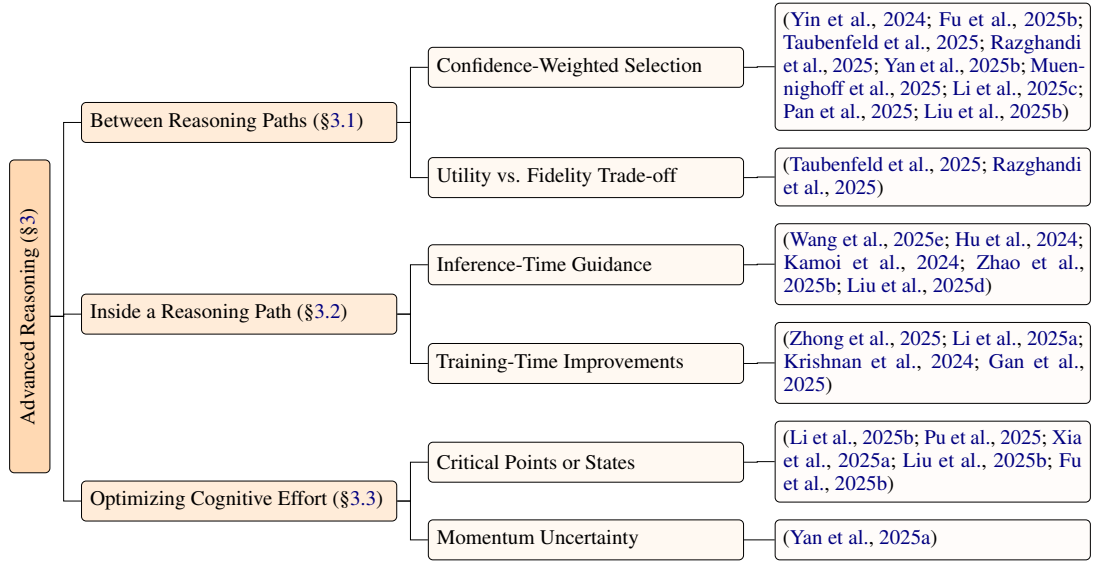

\end{document}